\documentclass[a4paper,conference]{IEEEtran}
\ifCLASSINFOpdf
\else
\fi
\newcommand{\icol}[1]{
  \begin{smallmatrix}#1\end{smallmatrix}%
}

\usepackage{times}

\usepackage[utf8]{inputenc} 
\usepackage[T1]{fontenc}    
\usepackage{hyperref}       
\usepackage{booktabs}       
\usepackage{amsfonts}       
\usepackage{nicefrac}       
\usepackage{microtype}      
\usepackage{amssymb}
\usepackage{stackengine}
\usepackage{makeidx}
\usepackage{color}
\usepackage{epsfig}
\usepackage{mathtools}
\usepackage{relsize}
\usepackage{multirow}
\usepackage{bm}

\usepackage{microtype}
\usepackage{subfigure}
\usepackage{booktabs} 
\usepackage{stackengine}

\usepackage{algorithm}
\usepackage{algpseudocode}

\algdef{SE}[SUBALG]{Indent}{EndIndent}{}{\algorithmicend\ }%
\algtext*{Indent}
\algtext*{EndIndent}

\definecolor{somegray}{rgb}{0.5, 0.5, 0.5}
\newcommand{\darkgrayed}[1]{\textcolor{somegray}{#1}}
\makeatletter
\newcommand*\titleheader[1]{\gdef\@titleheader{#1}}
\AtBeginDocument{%
  \let\st@red@title\@title
  \def\@title{%
    \vskip-1.6em
    \bgroup\normalfont\large\centering\@titleheader\par\egroup
    \vskip0.3em\st@red@title}
}
\makeatother
\titleheader{\darkgrayed{This paper has been accepted for publication at the \\
IAPR IEEE/Computer Society International Conference on Pattern Recognition (ICPR), Milan, 2021.}}

\title{
Unsupervised Feature Learning for Event Data: Direct vs Inverse Problem Formulation
}

\author{\IEEEauthorblockN{Dimche Kostadinov, Davide Scaramuzza}
\IEEEauthorblockA{Robotics and Perception Group \\
University of Zurich, Switzerland \\}
}

\begin{document}

\maketitle

\begin{abstract}
Event-based cameras record an asynchronous stream of per-pixel brightness changes. As such, they have numerous advantages over the standard frame-based cameras, including high temporal resolution, high dynamic range, and no motion blur. Due to the asynchronous nature, efficient learning of compact representation for event data is challenging. While it remains not explored the extent to which the spatial and temporal event "information" is useful for pattern recognition tasks. In this paper, we focus on single-layer architectures. We analyze the performance of two general problem formulations: the direct and the inverse, for unsupervised feature learning from local event data (local volumes of events described in space-time). We identify and show the main advantages of each approach. Theoretically, we analyze guarantees for an optimal solution, possibility for asynchronous, parallel parameter update, and the computational complexity. We present numerical experiments for object recognition. We evaluate the solution under the direct and the inverse problem and give a comparison with the state-of-the-art methods. Our empirical results highlight the advantages of both approaches for representation learning from event data. We show improvements of up to 9 $\%$ in the recognition accuracy compared to the state-of-the-art methods from the same class of methods.
\end{abstract}


%
\IEEEpeerreviewmaketitle

\section{Introduction}
By asynchronously capturing the light changes in a scene, the event-based camera offers an alternative approach for imaging, which is fundamentally different from the common frame-based cameras. Rather than measuring the “absolute” brightness at a constant rate, the event-based cameras measure per-pixel brightness changes (called “events”) in an asynchronous manner. Some of their main advantages are very high temporal resolution and low latency (both in the order of microseconds), very high dynamic range (140 dB vs. 60 dB of standard cameras), and low power consumption. Hence, event cameras have promising potential for pattern recognition, machine learning, computer vision, robotics, and other wearable applications in challenging scenarios (\textit{e.g.} high-speed motion and the scene has a high dynamic range). 

\begin{figure}[t!]
\centering
\begin{center}
\begin{minipage}[b]{1\linewidth}
\centering
\centerline{\includegraphics[width=\columnwidth]{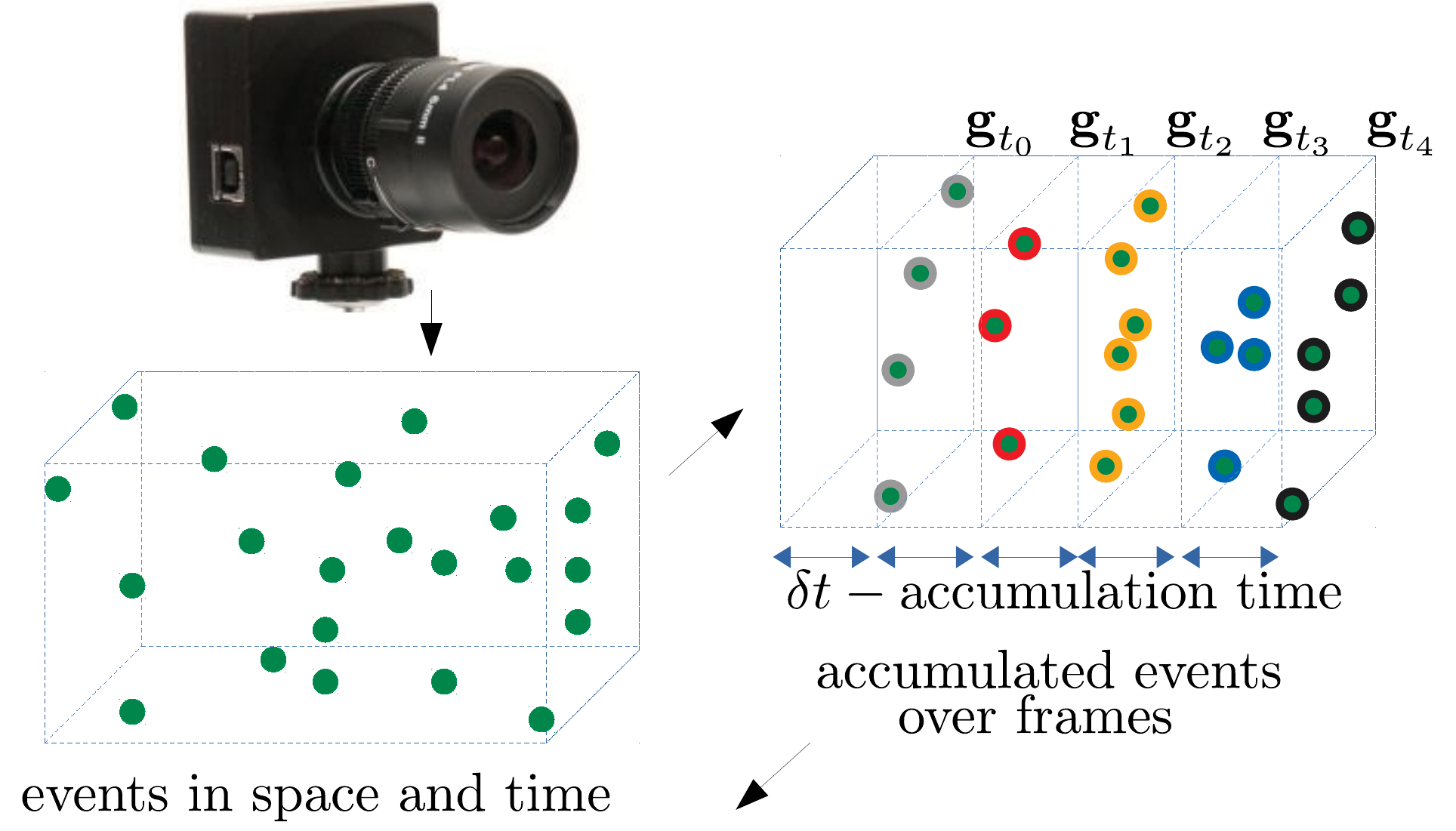}}
\end{minipage}
\begin{minipage}[b]{.9\linewidth}
\centering
\centerline{\includegraphics[width=\columnwidth]{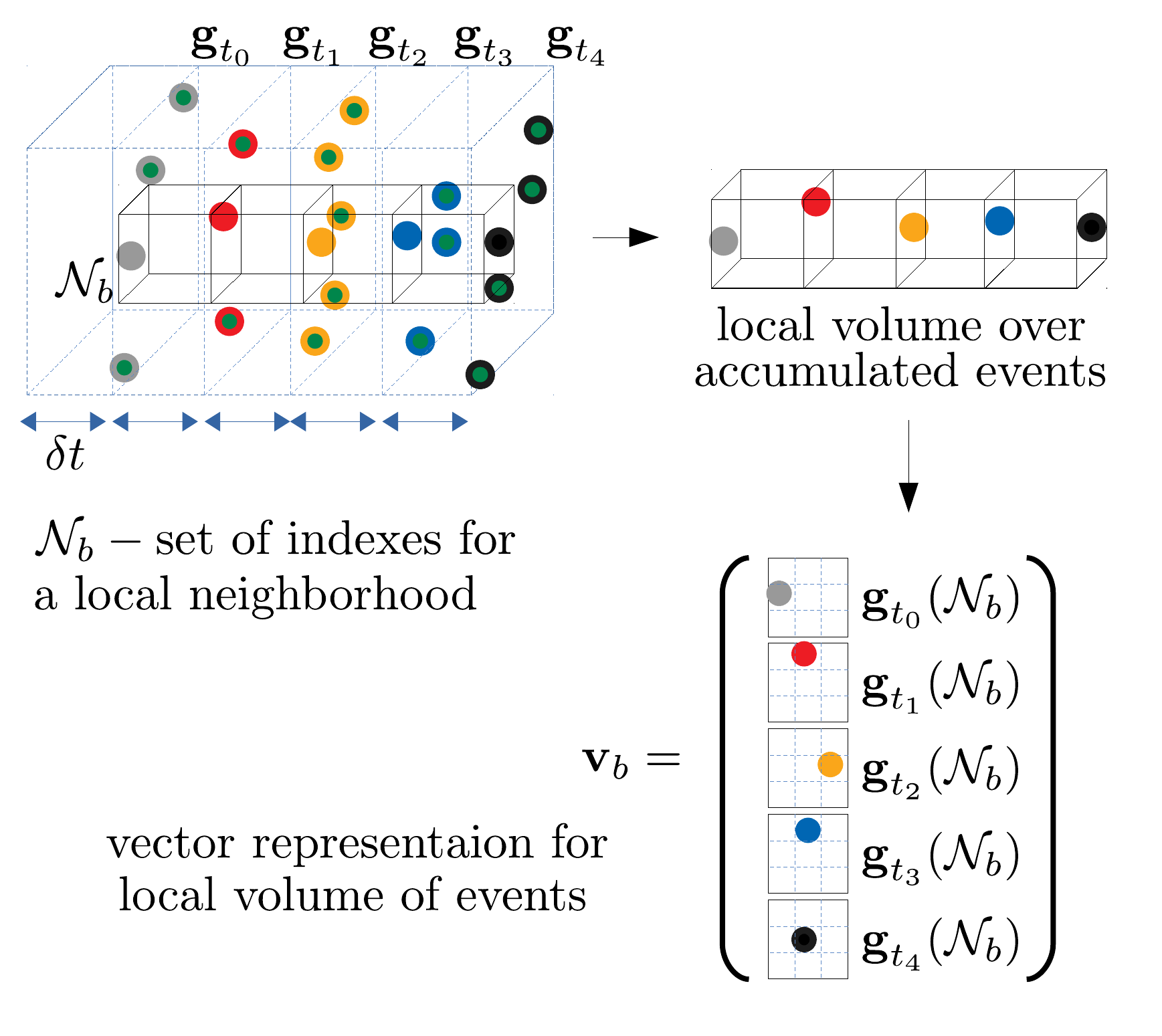}}
\end{minipage}
\vspace{-.1in}
\caption{An illustration for the construction of the local volume of aggregated events that we use as a vector input representation ${\bf v}_b$ in our unsupervised feature learning approach.}
\label{fig.lov.0}
\end{center}
\end{figure}

As data-driven sensors, the event-based camera output depends on the brightness change caused by the camera's motion or the objects' motion in the scene. The faster the motion, the more events per second are generated since each pixel adapts its sampling rate to the rate of change of the intensity signal that it monitors. 
One of the critical questions of the paradigm shift posed by event cameras is how to extract meaningful and useful information from the event stream to fulfill a given task. In the past, several unsupervised event-based features were proposed \cite{Guillermo:2019} for various tasks, like recognition, object detection, segmentation, and feature tracking. Based on how the feature is estimated, these methods can be grouped on two broad categories: (i) handcrafted and (ii) learned approaches. Concerning the used model, the latter category can be divided into two subgroups: (a) single-layer and (b) multi-layer architecture. 

Deep, multi-layered architectures have shown to be successful at many tasks,  but most of these methods process events in synchronous batches, sacrificing the asynchronous property of event data. Asynchronous and parallel learning for event-based data was addressed under the spiking neural networks (SNN) \cite{Dongsung:2018}. However,  SNNs were challenging to train due to the absence of an efficient equivalent to back-propagation learning method and the hyper-parameters' sensitivity. 
While the interpretations and the understanding of the learning dynamics, even for the most popular multi-layer architectures, remain challenging. On the other hand, data-adaptive, \textit{i.e.},  learned single-layer architectures for event-based data were not studied extensively. Moreover, the analysis for the appropriate problem formulation, which would be advantageous for efficient, asynchronous, and parallel learning from event-based data, was not fully explored.  As such,  it remains unknown to which extend a single-layer model could be useful for event-based data and how the spatial and temporal resolution of the event-based data impacts performance for a given task.

\begin{figure}[t!]
\centering
\begin{center}
\begin{minipage}[b]{.81\linewidth}
\centering
\centerline{\includegraphics[width=\columnwidth]{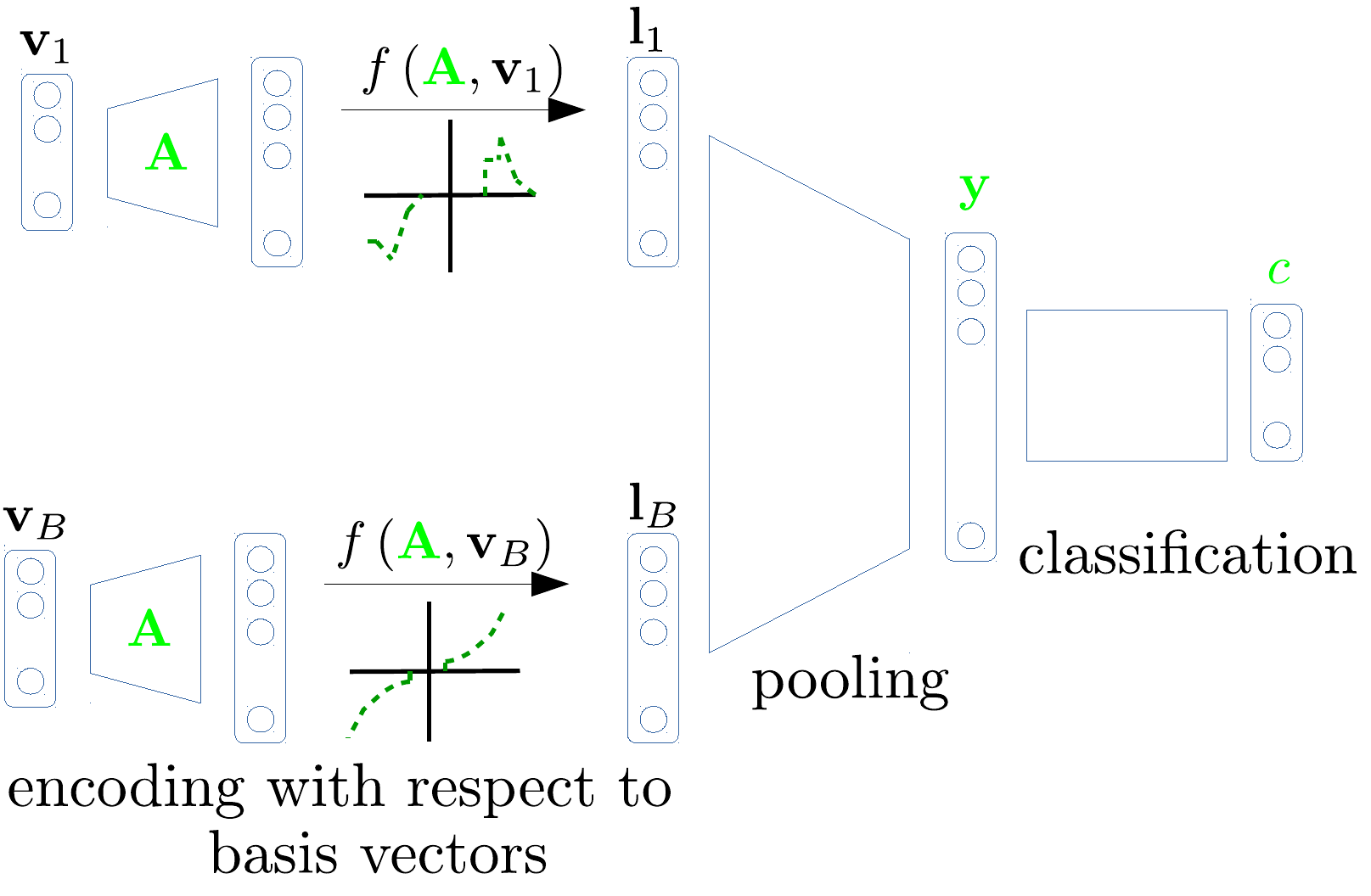}}
\end{minipage}
\caption{A schematic diagram which illustrates the main components in the proposed recognition pipeline.}
\label{fig.shem.diag}
\end{center}
\end{figure}


This paper analyzes two classes of single-layer methods for compact, information preserving, and task-relevant representation learning from event-based data. We focus on unsupervised learning of a set of basis vectors (or filter bank). We encode the event-based input data with respect to the basis vectors to produce features, which we use for recognition. In general, the problem of learning such basis vectors can be formulated as a direct \cite{DBLP:conf/icassp/RavishankarB14}, \cite{DBLP:journals/corr/RavishankarB15a}, \cite{Kostadinov2018:EUVIP} or inverse problem \cite{RubinsteinBE10}, \cite{LCKSVD:6516503} \cite{RubinsteinR:ADL}. 
\begin{itemize}
\item Under the inverse problem,  we would like to estimate (represent) the input event-data as a linear combination over a given set of basis vectors. 

\item  While under the direct problem, we would like to estimate (represent) the input event-data as a set of projections over a set of basis vectors.  
\end{itemize}
We highlight and show the advantages of both approaches. Theoretically, we reflect on the optimal solution and the complexity for asynchronous and parallel updates under both problem formulations. In both cases, we have to jointly estimate the representation and learn the set of basis vectors during training time. However, under the direct problem formulation, the complexity for estimating the representation is low. While under the inverse problem formulation, the complexity can be high, especially when the input event-based data dimension is high, or the number of basis vectors is high. We evaluate both methods on different data sets for the task of object recognition. Our validation shows improvements in the recognition accuracy while using event-data at low spatial resolution compared to the state-of-the-art methods.


\subsection{Contributions}
In the following, we give our main contributions.

\begin{itemize}
\item We analyze the recognition performance of a two part recognition pipeline: (i) unsupervised feature learning and (ii) supervised classifier over the encoded features. We show that with this simple single layer architecture we can achieve state-of-the-art result which outperforms handcrafted methods.


\item  In the unsupervised feature learning, we  address the direct and the inverse problem formulation, \textit{i.e.}, data adaptive basis vectors learning with respect to which we encode the event-based data. We investigate the solutions of the both problems, comment on the local optimal guarantees as well as highlight the complexity under asynchronous and parallel solution update. 

\item  We validate both approaches trough a numerical evaluation on different data sets for the task of object recognition.  
In addition, we also provide an analysis for different trade-offs and highlight the advantages of each approach. 
 We demonstrate that the direct problem formulation has identical recognition performance to the inverse problem formulation, but under the direct problem formulation, the learning complexity is lower. In addition to having local convergence guarantees, the direct problem formulation also has lower complexity for asynchronous and parallel update. Our numerical results, compared to the state-of-the-art methods from the same category show improvements of up to ${\bf 9} \%$ in the recognition accuracy. 
\end{itemize}

\section{Related Work}


Unsupervised feature learning is a well studied topic, \textit{e.g.}, \cite{PeterDiehl:2015, PiotrDollar:2015, Leutenegger:2011, Lowe:2004, Sivic:2009, PaulViola:2004} in the area of image processing, computer vision and machine learning for object recognition, detection and matching. On the other hand, only recently, the event-based vision has been used to address these problems. 

Analogously to the approaches used for a standard camera image, feature extraction approaches from event-based camera can be grouped into two broad categories handcrafted and learning-based. Spatio-temporal feature descriptors of the event stream were used for high-level applications like gesture recognition \cite{JunhaengLee:2012}, object recognition \cite{Paul:2015, HATS:2018} or face detection \cite{SouptikBarua:2016}. Low-level applications include optical flow prediction \cite{RyadBenosman:2012, RyadBenosman:2011} and image reconstruction \cite{PatrickBardow:2016}. As a data-driven model, asynchronous, spiking neural networks (SNNs) \cite{JunHaengLee:2016} have been applied to several tasks, \textit{e.g.}, object recognition \cite{JunHaengLee:2016, HFIST:2015, Jose:2013}, gesture classification \cite{ArnonAmir:2017}, and optical flow prediction \cite{RyadBenosman:2012}. However, computationally efficient equivalent to back-propagation algorithms still limits the usability of SNNs in complex real-world scenarios. Several works have recently proposed using standard learning architectures as an alternative to SNNs \cite{Ana:2018, DanielNeil:2016, HATS:2018, AlexZihaoZhu:2018}. Commonly, these works use a handcrafted event stream representation. Deep multi-layer architectures have shown to be successful at many tasks, but asynchronous and parallel learning and the interpretation and understanding of the learning dynamics remain challenging. While not much attention was given to single-layer architectures that exploit the appropriate problem formulation, which would be advantageous for efficient, asynchronous, and parallel learning from event-based data. 

\section{Paper Organization}

The rest of the paper is organized as follows. In Section \ref{sec:EventBasedCamera}, we introduce the working principle of the event-based camera. In Section \ref{sec:AsynchronousandParallelRepresentationLearningforEventBasedData}, we first present an overview of our approach, then we present how we form the input for the unsupervised learning algorithm. Afterward, we give the problem formulation for our unsupervised learning approach and describe our classifier. We devote Section \ref{sec:NumericalEvaluation} to numerical evaluation, while with Section \ref{sec:Conclusion}, we conclude the paper.

\section{Event Based Camera Working Principle}
\label{sec:EventBasedCamera}

In this section, we present the working principle of event cameras. 
The event-based cameras (like DVS [1]) at independent pixels locations measure “events” according to brightness change. Let $L([\icol{x \\ y}], t)= \log I([\icol{x \\ y}], t)$ be the logarithmic brightness at pixel location $\left[ \icol{x \\ y} \right]$ on the image plane. The event-based camera generates an event ${\bf e}_t=\{ [\icol{x \\ y}], t_k, p_k \}$ when the change in logarithmic brightness at pixel location $[\icol{x \\ y}]$ reaches a threshold $C$, \textit{i.e.}, 
\begin{equation}
\begin{aligned}
& \Delta L= L([\icol{x \\ y}], t_k) - L([\icol{x \\ y}], t - \Delta t) = p_t(x,y) C,
\label{WorkingPrincipleDescription:1}
\end{aligned}
\end{equation}
where $t$ is the time "stamp" of the event, $\Delta t$ is the time since the previous event at the same pixel location $[\icol{x \\ y}]$ and $p_t(x,y) \in \{ +1, -1 \}$ is the event polarity (i.e., sign of the brightness change). 
It is important to highlight that an event camera does not produce images at a constant rate, but rather a stream of asynchronous, sparse events in space and time. That is, depending on the visual input, the event-based camera, outputs data proportionally to the amount of brightness changes in the scene. 


\section{Unsupervised Representation Learning for Event Based Data}
\label{sec:AsynchronousandParallelRepresentationLearningforEventBasedData}

This section describes our approach, which focuses on learning features from local volumes of events. We adopt a multi-stage approach, which is similar to those employed in computer vision, as well as other feature learning works \cite{Coates:2011a}.

\subsection{Approach Overview}

Our approach consists of two parts (i) feature representation learning and (ii) classification. In Figure \ref{fig.shem.diag},  we show the schematic diagram, which illustrates the main components in the proposed recognition pipeline. The unsupervised learning part includes the following steps:
\begin{itemize}

\item[1)]  Extract random local volumes of events from unlabeled set of events for training and apply a pre-processing to the local volume of events.

\item[2)]  Learn a feature-mapping using an unsupervised
learning algorithm.

\end{itemize}
We address the problem of learning the feature mapping under unsupervised learning algorithm. We highlight that depending on the input representation, our single layer unsupervised feature learning approach has flexibility to accommodate and capture different features. That is, when the local spatial dimension equals to the spatial dimension of the event-based camera stream then our single layer architecture equals to a fully connected layer. In that case, our unsupervised learning of basis vectors consists of learning the weights in the equivalent fully connected layer. On the other hand, when the local spatial dimension is smaller to the spatial dimension of the event-based camera stream and the basis vectors are shared for all local volumes of accumulated events then our single layer architecture equals to a convolutional layer. Furthermore, if the length of the local volume of accumulated events equals to the total number of accumulation intervals then we have 2D convolution. This is indeed the case, since, we actually preform a convolution operation only over the spatial domain, \textit{i.e.}, the $X$ and $Y$ axis. Otherwise, when the length of the local volume of accumulated events is smaller to the total number of accumulation intervals, then we have additional temporal dimension over which we can preform the convolution operation and thus have a 3D convolution. In this paper, we focus on learning a basis vectors that correspond to learning a 2D convolutional filters. 

In the second part, given the learned feature mapping and a set of labels for the training events, we perform feature extraction and classification as follows:
\begin{itemize}

\item[1)]  Extract features from equally spaced local volumes of events covering the input event set and pool features together over regions of the input events to reduce the dimensionality of the feature vector.

\item[2)]  Train a linear classifier to predict the labels given
the feature vectors.

\end{itemize}

\subsection{Formation of Local Volumes of Events and Pre-processing}

Given an set of events across time as input, we extract and use local volumes of accumulated events. This representation is a vector ${\bf v}_b \in \Re^{B_x B_y T_l}$ that describes a local volume indexed by $b$, which has height $B_x $, width $B_y $ and length $T_l$ that equals to the number of time intervals over which we accumulate events. We construct ${\bf v}_b$ as follows:
\begin{equation}
\begin{aligned}
 {\bf v}_b = \left[ \icol{ {\bf g}_{t_l}( \mathcal{N}_b) \\ {\bf g}_{t_{l+1}}( \mathcal{N}_b) \\ . \\ . \\ . \\ {\bf g}_{t_{l+T_l}}( \mathcal{N}_b) } \right], \text{ where } {\bf g}_t = \sum_{t}^{t+\delta t} \left[ \icol{ p_{t}( {1,1} ) \\ p_{t}( {1,2} ) \\ . \\ . \\ . \\ p_{t}( {N_x,N_y} ) } \right], 
\label{WorkingPrincipleDescription:1.0.a}
\end{aligned}
\end{equation}
while $t \in \{0, ... ,T \}$ is the time index,  $\delta t$ is the duration of one accumulation interval, $\mathcal{N}_b$, $b \in \{1,...,B\}$ is the index for a spatial block with size $B_x \times B_y$, centered at spatial spatial position $(x, y)$. 
An illustration for the construction of the local volume of accumulated events is shown in Figure \ref{fig.lov.0}.

Our pre-processing consists of two parts. In the first part, we normalize each local volume of accumulated events ${\bf v}_b$ 
by subtracting the mean and dividing by the standard deviation of its elements. In fact, this corresponds to
normalization in the change of the event accumulation.  
After normalizing each input vector, in the second part, we whiten \cite{Coates:2011a} all ${\bf v}_b$ from the entire data set of events. This process is commonly used in the learning methods (e.g., \cite{Coates:2011a}) for standard images, but it is less frequently employed in pattern recognition from event-based data. 

\subsection{Unsupervised Feature Learning}
Our training data, \textit{i.e.}, ${\bf V}=[{\bf V}_1,...,{\bf V}_M]$ consists of set of vectors ${\bf V}_j=[{\bf v}_{j, 1},...,{\bf v}_{j, B}]$, where ${\bf v}_{j, b} \in \Re^{B_xB_x T_l}$ represents the local volume of temporally aggregated events over local spatial region indexed by $b$ for the event data $j$. We consider the direct and inverse problem formulations, where we jointly estimate the representations and learn a set  of  basis  vectors. In the following, we present the both problem formulations. 

\subsubsection{Inverse Problem Formulation}

The inverse problem formulation has the following from:
\begin{equation}
\begin{aligned}
\! \Big[ {\bf L}, {\bf D} \Big] \!  =\arg \min_{{\bf L}, {\bf D}}\frac{1}{2}\Vert{\bf V}-{\bf D}{\bf L} \Vert_F^2 \! + \! \lambda_0 m({\bf L} ) \! + \! \lambda_1\Omega({\bf D}),
\label{problem.formulation:inverse}
\end{aligned}
\end{equation}
where $\Vert . \Vert_F$ denotes Frobenius norm, $m({\bf L})=\sum_{j=1}^M \sum_{b=1}^B \Vert {\bf l}_{j,b} \Vert_1 $ and $\Omega({\bf D})=\lambda_2\Vert {\bf D} \Vert_2^2 - \lambda_3 \Vert {\bf D}{\bf D}^T-{\bf I} \Vert_2^2-\lambda_4 \log \vert \det {\bf D} {\bf D}^T \vert$ are constraints on the representations ${\bf L}=[{\bf L}_1,...,{\bf L}_M]$, ${\bf L}_j=[{\bf l}_{j,1},...,{\bf l}_{j,B}]$, ${\bf l}_{j,b} \in \Re^{K}$ and the dictionary (the set of basis vectors) ${\bf D}=[{\bf d}_1,...,{\bf d}_K]$, ${\bf d}_k \in \Re^{B_xB_x T_l}$ while $\lambda_0$ and $\lambda_1$ are Lagrangian parameters. Given the dictionary ${\bf D}$ under  the  inverse  problem \eqref{problem.formulation:inverse},  we  would  like  to  represent the local volume of events ${\bf v}_{j,b}$ as a sparse linear combination ${\bf l}_{j,b}$, \textit{i.e.}, ${\bf v}_{j,b}={\bf D}{\bf l}_{j,b}$. 

\subsubsection{Direct Problem Formulation}
The direct problem formulation has the following from:
\begin{equation}
\begin{aligned}
\Big[ {\bf L}, {\bf A} \Big]  \! =\arg \min_{{\bf L}, {\bf A}}\frac{1}{2}\Vert{\bf A}{\bf V}-{\bf L} \Vert_F^2 \! + \! \lambda_0 m({\bf L} ) \! + \!  \lambda_1 \Omega({\bf A} ),
\label{problem.formulation:direct}
\end{aligned}
\end{equation}
where ${\bf A}=\left[\icol{ {\bf a}_1^T \\ .\\. \\ {\bf a}_K^T }\right]$, ${\bf a}_k \in \Re^{B_xB_xT_l}$, while the constraints $m({\bf L})$ and ${\Omega}({\bf A})$ are equivalent with the ones defined for \eqref{problem.formulation:inverse}.  
Given the linear map ${\bf A}$, under the direct problem \eqref{problem.formulation:direct}, we would like to represent ${\bf v}_{j,b}$ as a two step nonlinear transform ${\bf l}_{l,b}=g({\bf A}{\bf v}_{l,b})$ consisting of: (i) linear mapping ${\bf A}{\bf v}_{j,b}$ and (ii) element-wise  nonlinearity $g({\bf A}{\bf v}_{l,b})$, which is induced by the constraint $m({\bf l}_{j,b})=\Vert {\bf l}_{j,b} \Vert_1$. 

Both, \eqref{problem.formulation:inverse} and \eqref{problem.formulation:direct} are non-convex in the variables $\{ {\bf L}, {\bf D} \}$ and $\{ {\bf L}, {\bf A} \}$, respectively. If the variable ${\bf D}$ in \eqref{problem.formulation:inverse} (or ${\bf A}$ in  \eqref{problem.formulation:direct}) is fixed, \eqref{problem.formulation:inverse} (or \eqref{problem.formulation:direct}) is convex, but if ${\bf L}$ is fixed, the reduced problem for  \eqref{problem.formulation:inverse} (or \eqref{problem.formulation:direct}) might not be convex due to the penalty function $\Omega$. 
Nonetheless, to solve \eqref{problem.formulation:inverse} (or  \eqref{problem.formulation:direct}) usually an iterative, alternating algorithm is used that has two steps: dictionary ${\bf D}$ (or transform ${\bf A}$) update and sparse coding.

\begin{table}[h]
\label{table.1}
\begin{center}
\begin{tabular}{|c|c|}
\hline 
Method & Local Convergence Guarantee \\
       & Under Sparsity Constraints \\
\hline 
Proposed (inverse) & exists  \\
Proposed (direct)  & exists \\
\hline
\end{tabular}
\end{center}
\caption{The local convergence guarantee.}
\end{table}

Considering the inverse problem \eqref{problem.formulation:inverse}, in the dictionary update step, given ${\bf L}$ that is estimated at iteration $t$, we use a K-SVD \cite{RubinsteinPE13}. In the sparse coding step, given ${\bf D}^{t+1}$, the sparse codes ${\bf l}_{l,b}^{t+1}$ are estimated using \cite{Tibshirani1994regressionshrinkage}. 

Considering the direct problem \eqref{problem.formulation:direct}, in the transform estimate step, given ${\bf L}$ that is estimated at iteration $t$, we use approximate closed form solution to estimate the transform matrix ${\bf A}^{t+1}$ at iteration $t + 1$. In the sparse coding step, given ${\bf A}^{t+1}$, the sparse codes ${\bf l}_{j,b}^{t+1}$ are estimated by a closed form solution.

\subsubsection{Local Convergence Guarantees}

Under sparsity constraints for the representations, and conditioning and coherence constraints \cite{Kostadinov2018:EUVIP} for the dictionary, a local convergence guarantee to both \eqref{problem.formulation:inverse} and \eqref{problem.formulation:direct} has been shown \cite{RubinsteinBE10}, \cite{RubinsteinR:ADL} and  \cite{Kostadinov2018:EUVIP}. However, it is important to note that under the direct problem formulation the complexity of the sparse coding step has very low computational complexity which is linear in the dimension of the representation ${\bf l}_{j,b}$, \textit{i.e.}, $O(B_x B_y T_l)$. While for the inverse problem the same complexity is higher \cite{DBLP:conf/icassp/RavishankarB13a}, \cite{DBLP:conf/icassp/RavishankarB14}. In addition, as advantage, the direct problem allows posing a class of penalties under which a low complexity closed form solution exists.   

\begin{table}[h]
\label{table.1}
\begin{center}
\begin{tabular}{|c|c|}
\hline 
Method & Complexity \\
& Under Sparsity Constraint \\
\hline 
Proposed (inverse) & $O(s B_x B_y T_l K)$  \\
Proposed (direct)  & $O(B_x B_y T_l)$ \\
\hline
Method & Complexity \\
& Under Broad Class of Constraint \\
\hline 
Proposed (inverse) & high  \\
Proposed (direct)  & low \\
\hline 
\end{tabular}
\end{center}
\caption{The complexity for estimating the unsupervised representation under sparsity constraints and under broad class of constraints.}
\end{table}

\subsection{Final Feature Composition and Classification}

Under either the direct or inverse problem formulation, we estimate basis vectors, which we consider as the parameters of a function that maps the input local volume of events to a new representation. We apply this mapping to our (labeled) training event-based data for classification.  

\subsubsection{Final Feature Composition}
We consider the learned basis vectors as the parameters of an encoding function $f : R^{B_x B_y T_l }
 \rightarrow R^K$, which represents our feature extractor. Many functions might be used to encode with respect to the learned basis vectors, here we use the triangle encoding, as presented by \cite{Coates:2011a}. Under our encoding function $f$, for any $B_x$-by-$B_y$-by-$T_l$ local volume of accumulated events ${\bf v}_{j, b}$, we 
compute the corresponding representation ${\bf l}_{j, b} \in R^
K$. Moreover, we define a (single layer) representation for the set of events by applying the function $f$ to all of the local volumes of accumulated events. That is, given the set of events defined over $N_x$-by-$N_y$-by-$T$ volume of event locations, for each of the local volumes ${\bf v}_{j, b}$ described by the spatial index $b \in \mathcal{N}_b$,  we compute the representation ${\bf l}_{j,b}$. More formally, we let ${\bf l}_{j,b}$ to be the $K$-dimensional representation extracted for location index $b$, from the input set of events indexed by $j$. 

We reduce the dimensionality of the event-based data representation by pooling, which is similar as proposed by \cite{Coates:2011a}, but instead of image patches, we operate on local volumes of accumulated events and construct the final representation $ {\bf y}_j = \left[  \icol{ \sum_{b_1} {\bf l}_{j, b_1} \\  . \\ .\\ . \\ \sum_{b_4} {\bf l}_{j, b_4}  } \right] \in \Re^{4K}$.

\subsubsection{Classification}  

In our classification, we use the pooled feature vectors ${\bf y}_{j}$ for each training event-based data and its corresponding label. We apply (L2) SVM \cite{Vapnik:1995:SL:ER} classification, with regularization parameter that is determined by cross-validation.

\subsection{Asynchronous and Parallel Update in Feature Learning}
We note that the differences between the inverse problem \eqref{problem.formulation:inverse} and the direct problem \eqref{problem.formulation:direct} emerge only if the set of basis vectors is over-complete or under-complete. Since, under an orthonormal set of basis vectors the two problems are equivalent. Considering an over-complete set of basis vectors, the solution in the sub-problems related to both the inverse problem \eqref{problem.formulation:inverse} and the direct problem \eqref{problem.formulation:direct} have major impact on the possibility for asynchronous and parallel update for the proposed unsupervised feature learning approach.  

During the sparse coding step for the inverse problem formulation \eqref{problem.formulation:inverse}, at any change (even small) of the input representation, a solution to inverse problem has to be estimated, which would lead to high computational complexity and challenges in parallel update of the representation. In contrast, the same step in the direct problem formulation \eqref{problem.formulation:direct} has a closed from solution. Moreover, under \eqref{problem.formulation:direct}, it is straightforward to update each element of the representation in parallel and in an asynchronous fashion, with no additional increase in the computational complexity. 

During the basis set update, under the conditioning and coherence constraints, for both the inverse \eqref{problem.formulation:inverse} and the direct \eqref{problem.formulation:direct} problem formulation, parallel update is challenging. Additional, structure enforcing constraint on the basis set might be helpfully towards parallel and asynchronous update of the basis set, especially under the direct \eqref{problem.formulation:direct} problem formulation.


\section{Numerical Evaluation}
\label{sec:NumericalEvaluation}

In this section, we evaluate the potential of our approach and provide comparative results between our algorithm and the state-of-the-art methods. We consider the task of object recognition over three publicly available data sets. In the following subsection, we describe the setup for the preformed experiments and finally present and discuss the results.

\subsection{Data Sets, Setup and Analysis}
In our evaluation, we used N-MNIST  \cite{N-MNIST:N-CALTECH}, N-Caltech101  \cite{N-MNIST:N-CALTECH} and N-Cars \cite{N-CARS} event-based data sets. We use our classification pipeline, which we presented in the earlier section. We learned the over-complete basis set under the direct and the inverse problem formulation. As an encoding function, we use the triangle encoding as proposed in \cite{coates2011importance, Coates:2011a}, by which we encode the local volume of events with respect to the learned set of basis vectors. After the puling stage over all the encoded representations for the  local volumes of events, we use the final representation to learn a linear SVM classifier.

Considering the N-MNIST data set, we use the final representations from the training data set in order to learning the classifier, while we use the final representations from the test data set for evaluation of the recognition accuracy. Considering the N-Caltheh101 and N-Cars data sets, we use $80 \%$ of the data set for training and the remaining to test the recognition performance.

We compare the results of our approach under the direct and the inverse problem formulation for learning the basis set with the state-of-the-art methods. Our evaluation includes comparison to the methods under single layer as well as multi-layer architectures which use event-based data. 
In addition, we analyze the impact on the recognition performance when we change different parameters in the components of our pipeline. We present the recognition results under varying  (i) number of basis vectors, (ii) size of the local volume and (iii) number of accumulation bins.

\begin{table}[h]
\label{table.1}
\begin{center}
\begin{tabular}{|c|c|}
\hline 
Method & Acc. $\%$ \\
\hline 
Hfist \cite{Hfist} & 06.0 \\
HOTS \cite{HOTS}  & 21.0 \\
Garbor-SNN \cite{Gabor:SNN} & 19.2 \\
HATS \cite{HATS:2018}  & 64.2 \\
DART \cite{DART}  & 70.3 \\
\hline 
\end{tabular}
\begin{tabular}{|c|c|}
\hline 
Method & Acc. $\%$ \\
\hline 
EST \cite{EST:2019} & 81.7 \\
VID2E \cite{VID2E:2020} & 90.1 \\
\hline  
\end{tabular}

\begin{tabular}{cc}
&
\end{tabular}
\end{center}

\begin{center}
\begin{tabular}{|c|c|}
\hline 
Method & Acc. $\%$ \\
\hline 
Proposed (inverse) & \textbf{78.4} \\
Proposed (direct)  & \textbf{77.1} \\
\hline
\end{tabular}
\end{center}
\caption{A comparison of the average precision accuracy for our approach and the state-of-the-art single layer and multi-layer methods, which use event data as input.}
\label{table.avg.acc.cmp.all}
\begin{center}
\begin{tabular}{|c|c|c|}
\hline 
Data set & Acc. $\%$ (inverse) & Acc. $\%$ (direct)  \\
\hline 
N-MNIST  &98.1 & 96.8 \\
N-Calteh101 & 78.4 & 77.1 \\
N-CARS  & 84.7 & 81.3   \\
\hline
\end{tabular}
\end{center}
\caption{The average precision accuracy of the proposed method under direct and inverse problem formulation for unsupervised feature learning over the used N-MNIST, N-Calteh101 and N-CARS event-base data sets.}
\label{table.avg.acc.across.datasets}
\end{table}

\begin{table}[h!]
\label{table.1}
\begin{center}
\begin{tabular}{|c|c|c|c|c|c|}
\hline 
Method & \multicolumn{4}{c|}{Number of Basis Vectors}  \\
\hline 
& 1000 & 1500 & 1700 & 2000 \\
\hline 
Proposed (inverse)  & 73.2 & 74.5  & 78.4 & 76.0 \\
Proposed (direct)   & 74.3& 77.0& 77.1 &75.5  \\
\hline
\end{tabular}
\end{center}
\begin{center}
\begin{tabular}{|c|c|c|c|c|c|}
\hline 
Method & \multicolumn{4}{c|}{Size of the Local Volume}  \\
\hline 
& 4$\times$4$\times$4 & 4$\times$12$\times$12 & 4$\times$16$\times$16 & 4$\times$21$\times$ 21 \\
\hline 
Proposed (inverse) & 69.6 & 78.4& 76.4 & 75.2   \\
Proposed (direct)  & 64.8& 77.1& 74.5 &75.1    \\
\hline
\end{tabular}
\end{center}
\label{table.1}
\begin{center}
\begin{tabular}{|c|c|c|c|c|c|}
\hline 
Method & \multicolumn{4}{c|}{Number of Accumulation Intervals}  \\
\hline 
& 2 & 4 & 7 & 10 \\
\hline 
Proposed (inverse) & 61.7 & 72.4 & 78.4 & 76.3   \\
Proposed (direct)  & 63.2&  69.1& 77.1 & 74.1    \\
\hline
\end{tabular}
\end{center} 
\caption{The average precision accuracy under varying: (i) number of basis vectors, (ii) size of the local volume and (iii)
number of accumulation intervals.}
\vspace{-.1in}
\label{table.varying.acc.intervals}
\end{table}

\begin{figure*}[t!]
\centering
\begin{center}
\begin{minipage}[b]{.25\linewidth}
\centering
\centerline{\includegraphics[width=\columnwidth]{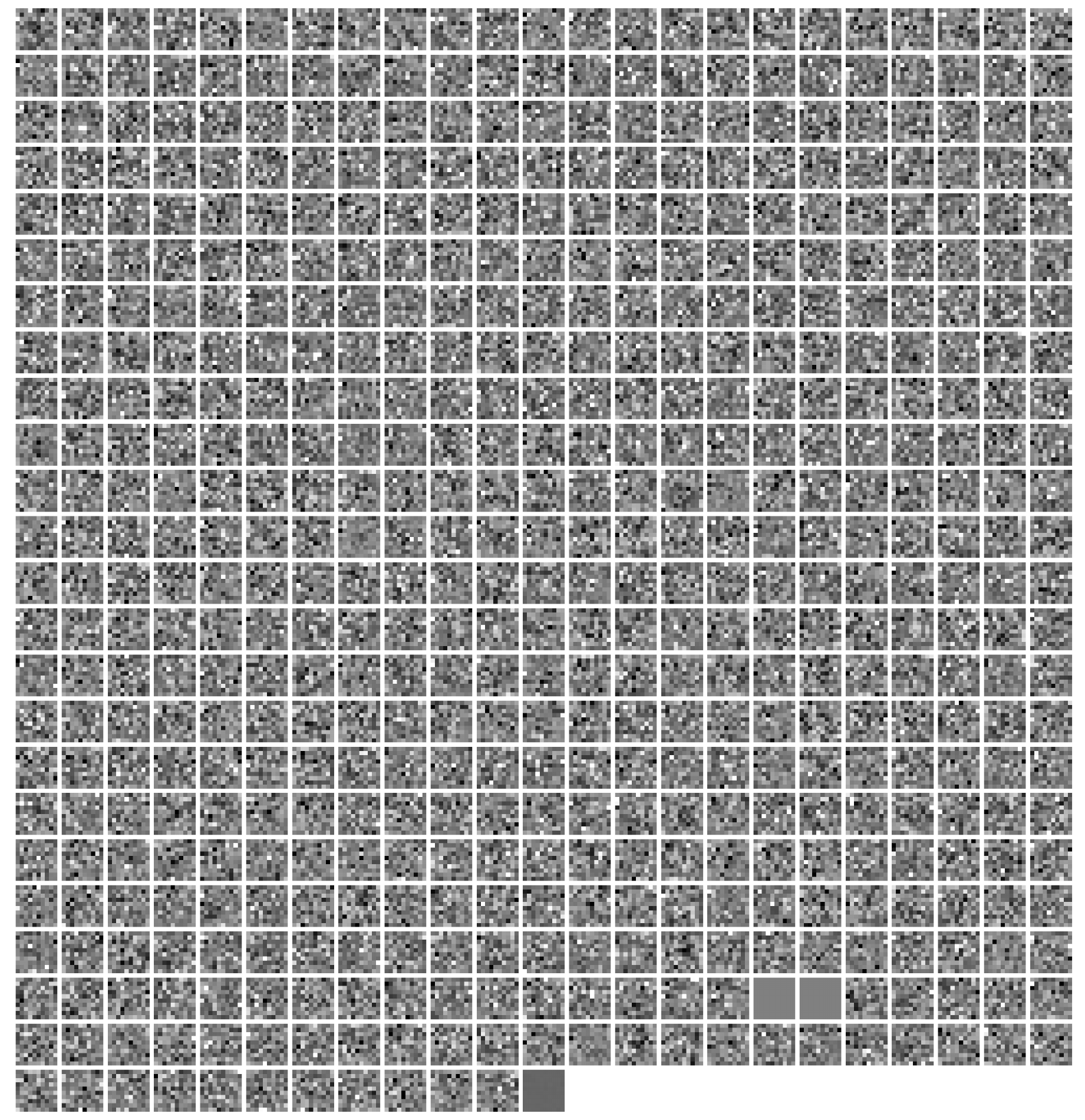}}
\end{minipage}
\begin{minipage}[b]{.2\linewidth}
\centering
\text{ }
\end{minipage}
\begin{minipage}[b]{.25\linewidth}
\centering
\centerline{\includegraphics[width=\columnwidth]{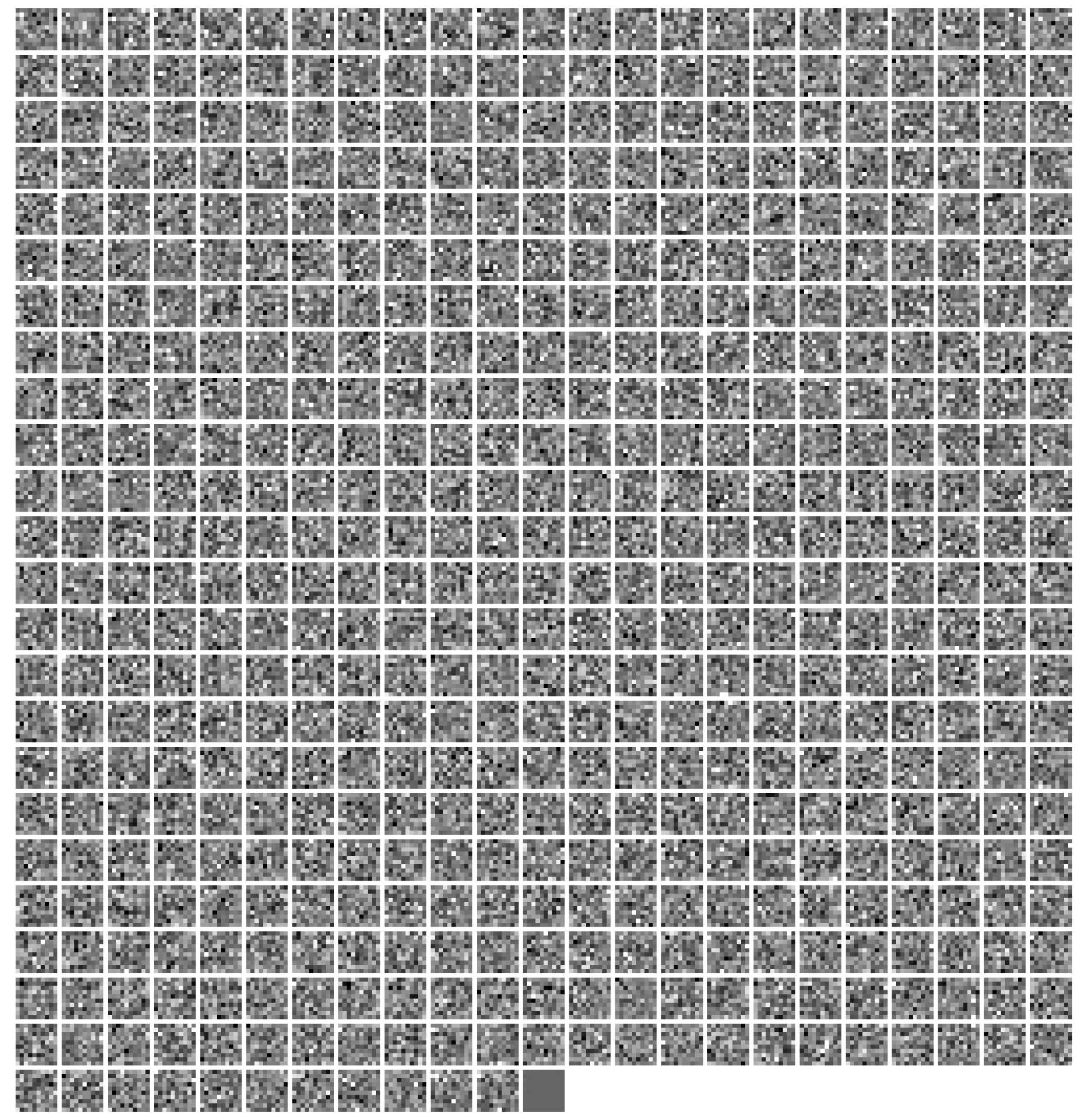}}
\end{minipage}
\caption{Visualization for a subset of the learned basis vectors under the inverse and direct problem formulation for the N-CARS data set.}
\label{fig.basis.set}
\end{center}
\vspace{-.2in}
\end{figure*}


\subsection{Result Discussion}

The results of our numerical evaluation are shown in Tables \ref{table.avg.acc.cmp.all},
\ref{table.avg.acc.across.datasets}, 
and \ref{table.varying.acc.intervals}. 

In Tables \ref{table.avg.acc.cmp.all} and \ref{table.avg.acc.across.datasets}, we show a comparison between the average recognition accuracy of our approach under direct and inverse problem formulation and the state-of-the-art single layer methods which use as input event-based data and frame based data. As we can observe, highest accuracy achieves the proposed unsupervised learning approach under inverse problem formulation. The unsupervised learning approach under the direct problem formulation also has high accuracy, which is a bit lower, but the learning complexity is also lower compared to the unsupervised learning approach under inverse problem formulation. Moreover, we can see that the proposed unsupervised learning approach outperforms the comparing state-of-the-art methods from the same category. It is important to note that under the proposed unsupervised learning approach, we use as input an event data that has $\times 4$ lower spatial resolution. That is, we use $\sim {\bf 1}$ order of magnitude less event data compared to the state-of-the-art methods. 

In Table \ref{table.avg.acc.cmp.all}, we also show a comparison between the average recognition accuracy of our approach under direct and inverse problem formulation and the state-of-the-art multi-layer methods which use as input event-based data. We can see that compared to the EST \cite{EST:2019} method, the accuracy is lower, but only about $3.3 \%$. On the other had, compared to the VID2E \cite{VID2E:2020} multi-layer method, we have bigger gap in the achieved accuracy. However, we have to note that under the multi-layered methods the authors used (i) pre-trained networks (using image-net \cite{imagenet:2009} data set), (ii) additionally simulated (artificially generated) and augmented event-data, while in our approach, we did not used any additionally simulated or external data. The recognition accuracy over each of the event-based data set is shown in Table \ref{table.avg.acc.across.datasets} and we can see that the recognition accuracy is high. 

In Table 
\ref{table.varying.acc.intervals},  we  show the recognition results under varying different  parameters in  the components  of  our  recognition pipeline.  As we can see the higher number of basis vectors we have higher recognition accuracy, but up to a certain point. We explain this by the fact that to small number of overcomplete basis vectors does not capture well the inherent invariances over all of the local volumes of events, while to big number of overcomplete set of basis vectors reduces the discrimination property. Interestingly, we have similar behavior when we vary the size of the local volume. However, we explain this by the fact that under very small local volumes the space of variation is to small, hence lower distinguishability between the local blocks, which results in low recognition accuracy. On the other hand, under very big local volumes the space of variation is big, which would imply that we have to use a bigger number of basis vectors\footnote{In Fig. \ref{fig.basis.set}, we give visualization of a subset of the learned basis vectors for the N-Cars event-based data set under the inverse and direct problem formulation. We note that there is not much of visual difference between the basis  vectors when we have regularization for the conditioning and the coherence of the basis set.}. 

Considering the impact of the number of accumulation intervals, we note that we achieve the highest accuracy when we have $7$ accumulation intervals. We suspect that the reason for low accuracy at too small or too big numbers of accumulation intervals is due to the fact that our events are triggered by motion, which we do not know in advance. This adds additional variability in the distribution of the events across the spatial domain that can be interpreted as noise corruption.

\section{Conclusion}
\label{sec:Conclusion}

In this paper,  we presented an analysis of the performance of two general problem formulations, \textit{i.e.}, the direct and the inverse, for unsupervised feature learning from local event data. We identified and show the main advantages of each approach theoretically and by a numerical validation. 

Empirically, we demonstrated that the direct problem formulation has similar recognition performance to the inverse problem formulation,  but the learning complexity is lower under the direct problem formulation.  The direct solution to the problem formulation has lower complexity, advantageous for an asynchronous and parallel update. In our numerical evaluation, the comparison with the state-of-the-art methods from the same category showed improvements of up to $\bf 9 \%$ in the recognition accuracy while using fewer event data. 

We leave the analysis of different encoding functions, the joint learning of  features and classifiers, and the analysis for a multi-layered unsupervised feature learning as feature work.

\bibliography{icpr2020}
\bibliographystyle{plain}

\end{document}